\documentclass{article}

\usepackage{tikz}
\usepackage{annotate-equations}
\usepackage{tablefootnote}

\usepackage{amsmath}
\usepackage{amssymb}
\usepackage{mathtools}
\usepackage{hyperref}
\usepackage{cleveref}
\usepackage{xcolor}

\usepackage[nonatbib,preprint]{arxiv}
\usepackage{wrapfig}

\usepackage[utf8]{inputenc} %
\usepackage[T1]{fontenc}    %
\usepackage{url}            %
\usepackage{booktabs}       %
\usepackage{amsfonts}       %
\usepackage{mathtools}
\usepackage{enumerate}
\usepackage{nicefrac}       %
\usepackage{microtype}      %
\usepackage{subcaption}
\usepackage{multirow} %
\usepackage{caption} %
\usepackage{hhline} %

\usepackage{arydshln}
\usepackage{enumitem}
\usepackage[normalem]{ulem}

\usepackage{tabu}
\usepackage[edges]{forest}
\usepackage{graphicx}
\setitemize{noitemsep,topsep=0.pt,parsep=0pt,partopsep=0pt}
\usepackage{authblk}

\usepackage{colortbl}    

\usepackage{amsmath,amsfonts,bm}

\def\eqref#1{equation~\ref{#1}}

\def\1{\bm{1}}

\DeclareMathAlphabet{\mathsfit}{\encodingdefault}{\sfdefault}{m}{sl}
\SetMathAlphabet{\mathsfit}{bold}{\encodingdefault}{\sfdefault}{bx}{n}

\DeclareMathOperator*{\argmin}{arg\,min}

\newcommand{\cb}{{\boldsymbol c}}

\newcommand{\s}{{\boldsymbol s}}

\newcommand{\n}{{\boldsymbol n}}

\newcommand{\x}{{\boldsymbol x}}
\newcommand{\y}{{\boldsymbol y}}

\newcommand{\epsilonb}{{\boldsymbol \epsilon}}

\newcommand{\Ed}{{\mathbb E}}

\newcommand{\Cd}{{\mathbb C}}

\newcommand{\Nc}{{\mathcal N}}

\newcommand{\code}[1] {\texttt{#1}}

\usepackage{capt-of}
\usepackage{diagbox}

\definecolor{C0}{rgb}{0.121569, 0.466667, 0.705882}
\definecolor{C1}{rgb}{1.000000, 0.498039, 0.054902}
\definecolor{C2}{rgb}{0.172549, 0.627451, 0.172549}
\definecolor{C3}{rgb}{0.839216, 0.152941, 0.156863}
\definecolor{C4}{rgb}{0.580392, 0.403922, 0.741176}
\definecolor{C5}{rgb}{0.549020, 0.337255, 0.294118}
\definecolor{C6}{rgb}{0.890196, 0.466667, 0.760784}
\definecolor{C7}{rgb}{0.498039, 0.498039, 0.498039}
\definecolor{C8}{rgb}{0.737255, 0.741176, 0.133333}
\definecolor{C9}{rgb}{0.090196, 0.745098, 0.811765}
\definecolor{trolleygrey}{rgb}{0.5, 0.5, 0.5}

\definecolor{BrickRed}{rgb}{0.6,0,0}
\definecolor{RoyalBlue}{rgb}{0,0,0.8}
\definecolor{Tdgreen}{rgb}{0,0.4,0.7}
\definecolor{pinegreen}{rgb}{0.0, 0.47, 0.44}
\definecolor{cornellred}{rgb}{0.7, 0.11, 0.11}
\definecolor{cadmiumgreen}{rgb}{0.0, 0.42, 0.24}
\definecolor{spirodiscoball}{rgb}{0.06, 0.75, 0.99}
\definecolor{mylightblue}{rgb}{0.85, 0.90, 0.94}
\definecolor{maroon}{cmyk}{0,0.87,0.68,0.32}

\usepackage{pifont}
\definecolor{c0}{rgb}{0.392, 0.051, 0.373}
\definecolor{c1}{rgb}{0.851, 0.086, 0.337}

\newcommand{\method}{ContextMRI}

\colorlet{lightc0}{c0!10}
\colorlet{lightc1}{c1!10}
\newcommand{\stddev}[1]{{\scriptsize \textcolor{gray}{$\pm$ #1}}}
\def\eqref#1{Eq.~\ref{#1}}


\usepackage[style=numeric, maxbibnames=15, maxcitenames=3, natbib=true,maxalphanames=10, backend=biber, sorting=nty, backref=true]{biblatex}
\DefineBibliographyStrings{english}{%
  backrefpage = {page},%
  backrefpages = {pages},%
}

\newcommand{\contextMRI}{ContextMRI} 

\title{\contextMRI: Enhancing Compressed Sensing MRI through Metadata Conditioning}

\author{
    Hyungjin Chung\textsuperscript{1,*}, 
    Dohun Lee\textsuperscript{1,*}, 
    Zihui Wu\textsuperscript{2}, \\
    Byung-Hoon Kim\textsuperscript{3}, 
    Katherine L. Bouman\textsuperscript{2},
    Jong Chul Ye\textsuperscript{1}
}

\affil{
    \textsuperscript{1}KAIST \quad
    \textsuperscript{2}Caltech \quad
    \textsuperscript{3}Yonsei University, College of Medicine \quad
}

\begin{document}

\maketitle

\begin{abstract}
Compressed sensing MRI seeks to accelerate  MRI acquisition processes by sampling fewer $k$-space measurements and then reconstructing the missing data algorithmically. The success of these approaches often relies on strong priors or learned statistical models. While recent diffusion model-based priors have shown great potential, previous methods typically ignore clinically available metadata (e.g. patient demographics, imaging parameters, slice-specific information). In practice, metadata contains meaningful cues about the anatomy and acquisition protocol, suggesting it could further constrain the reconstruction problem. In this work, we propose ContextMRI, a text-conditioned diffusion model for MRI that integrates granular metadata into the reconstruction process. We train a pixel-space diffusion model directly on minimally processed, complex-valued MRI images. During inference, metadata is converted into a structured text prompt and fed to the model via CLIP text embeddings. By conditioning the prior on metadata, we unlock more accurate reconstructions and show consistent gains across multiple datasets, acceleration factors, and undersampling patterns. Our experiments demonstrate that increasing the fidelity of metadata—ranging from slice location and contrast to patient age, sex, and pathology—systematically boosts reconstruction performance. This work highlights the untapped potential of leveraging clinical context for inverse problems and opens a new direction for metadata-driven MRI reconstruction.
Code is available at \href{https://github.com/DoHunLee1/ContextMRI}{https://github.com/DoHunLee1/ContextMRI}
\end{abstract}


\begin{figure}[!t]
    \centering
    \includegraphics[width=\linewidth]{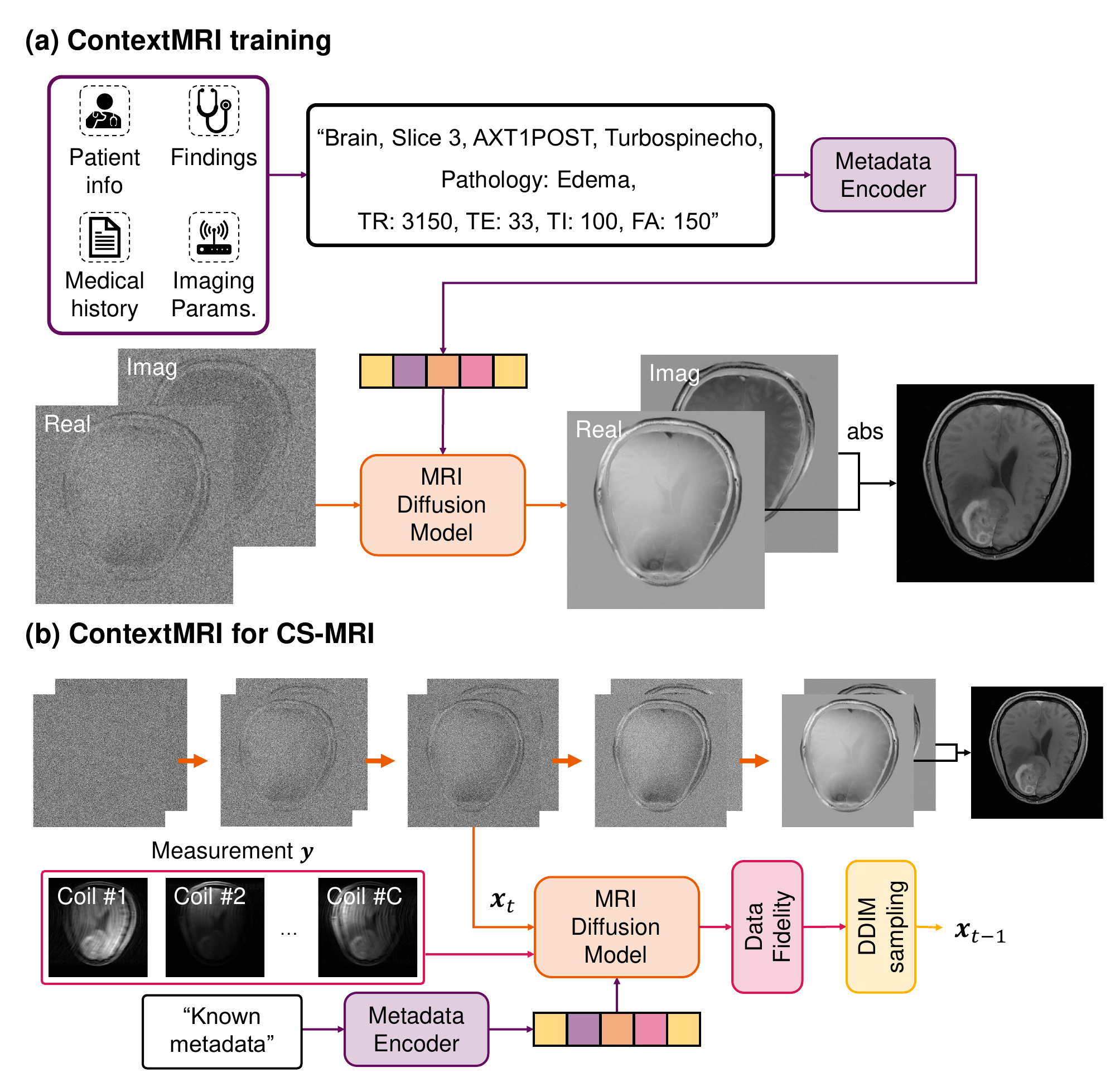}
    \caption{\bf\footnotesize Illustration of the method used in \method. (a) We convert available metadata into text format, which is encoded as a feature vector as an additional input to the diffusion model. The diffusion model is trained in pixel space with MVUE complex-valued images. (b) \method~can be used for CS-MRI by leveraging off-the-shelf diffusion model-based inverse problem solvers while additionally incorporating available metadata.}
    \label{fig:\method}
\end{figure}

\begin{figure}[!th]
    \centering
    \begin{minipage}[b]{0.39\textwidth}
        \centering
        \includegraphics[width=\textwidth]{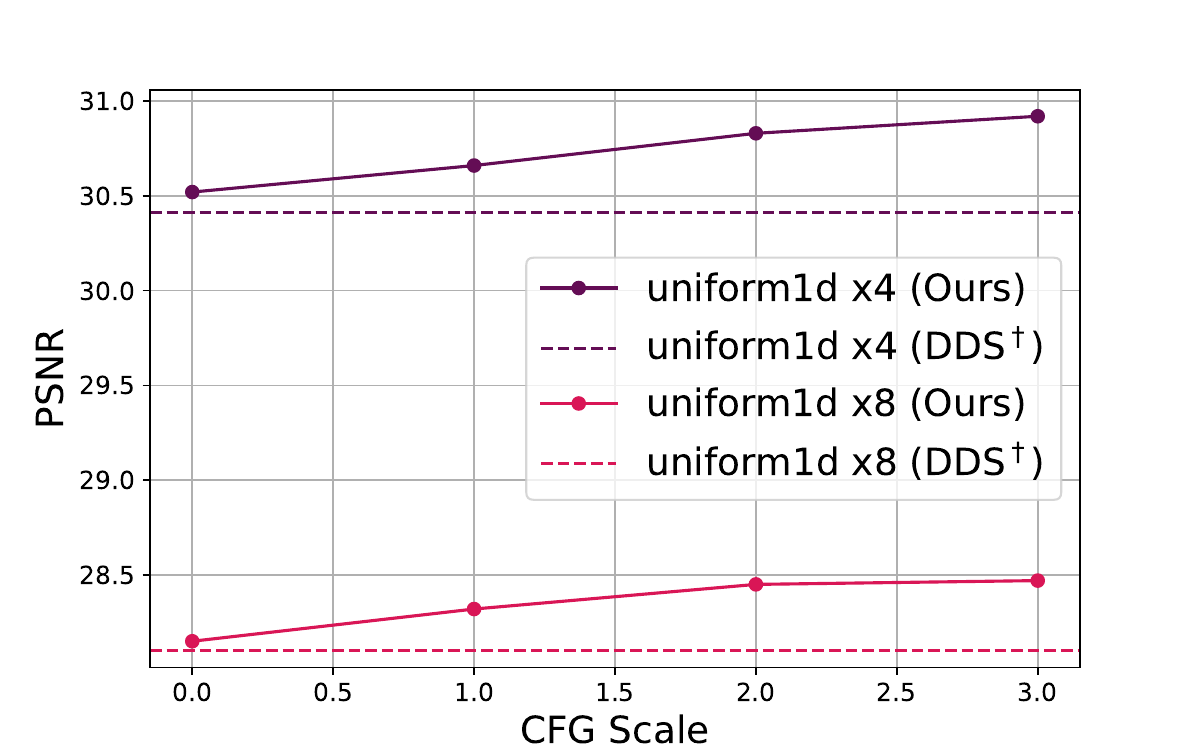}
        \subcaption{\bf\footnotesize CFG vs. PSNR on fastMRI knee}
        \label{subfig:knee}
        \includegraphics[width=\textwidth]{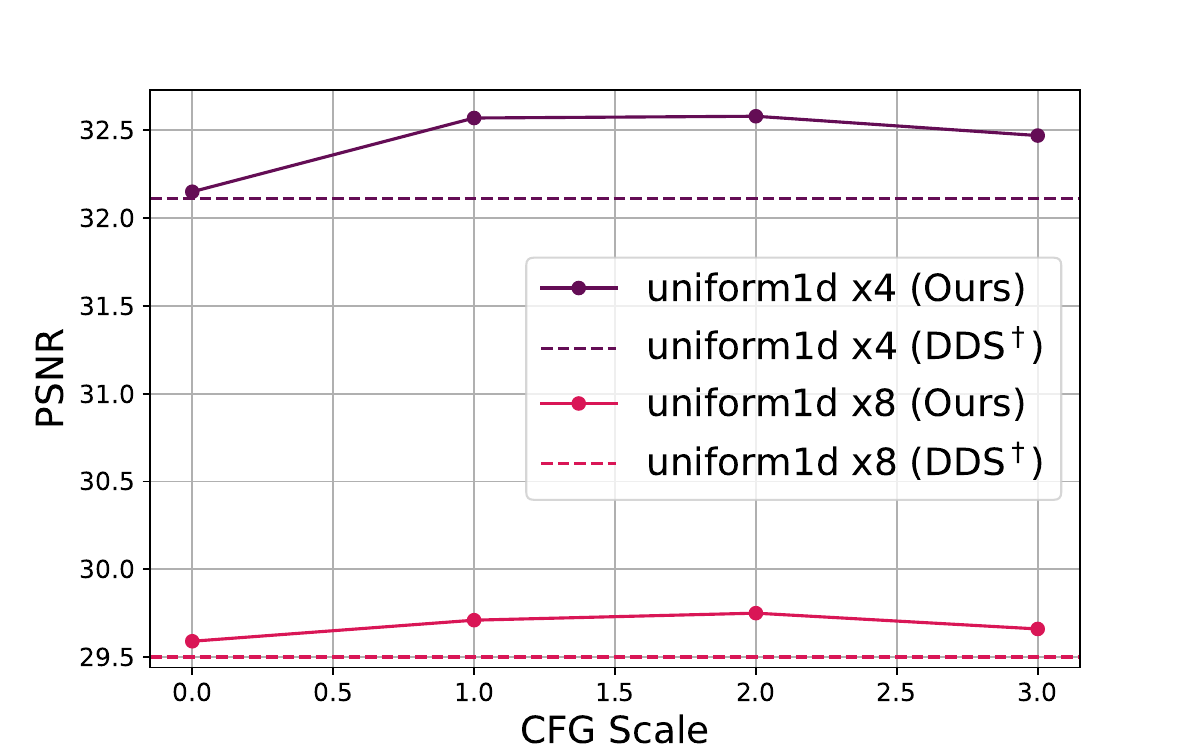}
        \subcaption{\bf\footnotesize CFG vs. PSNR on fastMRI brain}
        \label{subfig:brain}
    \end{minipage}%
    \hfill
    \begin{minipage}[b]{0.61\textwidth}
        \centering
        \includegraphics[width=\textwidth]{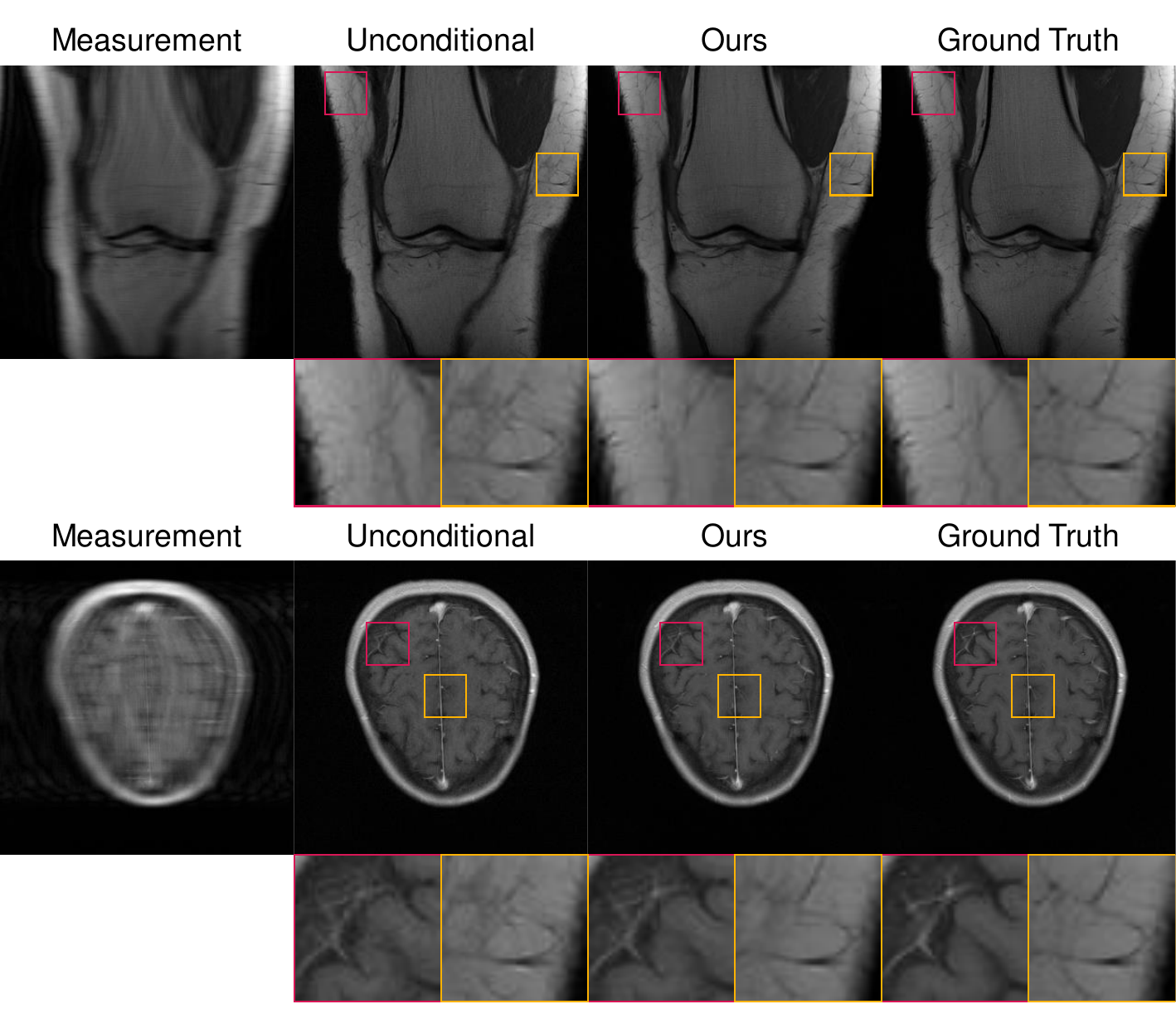}
        \subcaption{\bf\footnotesize Qualitative comparison of \method~against unconditional reconstruction}
        \label{subfig:main_results}
    \end{minipage}

    \caption{\bf\footnotesize Quantitative and qualitative comparison of the proposed method against unconditional reconstruction (DDS). The dashed line with $\dagger$ indicates that DDS was performed with the diffusion model used in the original work of \cite{chung2024decomposed}, which is the reason for the difference in performance between the unconditional version of our method and the dotted line.}
    \label{fig:main_results}
\end{figure}


\section{Introduction}
\label{sec:intro}

Diffusion models~\cite{song2020score,ho2020denoising} have recently emerged as pivotal tools for solving inverse problems in medical imaging~\cite{chung2022score,jalal2021robust,song2022solving}. These methods, referred to as diffusion model-based inverse problem solvers (DIS)~\cite{kadkhodaie2021stochastic,kawar2022denoising,chung2023diffusion}, aim to estimate the posterior distribution of the signal $\x$ given measurements $\y$, rather than relying on direct regression from undersampled data to fully sampled reconstructions. By leveraging pre-trained diffusion models as universal priors, DIS approaches can flexibly adapt to different imaging physics (e.g. changes in coil sensitivities, sampling patterns), while providing high-quality reconstruction images in line with Bayesian posterior inference.

Most modern diffusion models trained on natural images are conditional, with text being the most common conditioning ~\cite{rombach2022high,saharia2022photorealistic,videoworldsimulators2024}. Training a conditional model provides at least two advantages. First, it enhances controllability with techniques such as classifier-free guidance (CFG)~\cite{ho2021classifierfree}. Second, it was shown that it is significantly easier to train a conditional diffusion model than an unconditional one on the same dataset~\cite{peebles2023scalable}, as text signals can be a useful signal for disentangling the representation space. 

In terms of image restoration through DIS, several recent works~\cite{chung2024prompttuning,kim2023regularization,kim2025dreamsampler} have shown the promise of leveraging text conditions. It was shown that one can achieve better reconstruction performance by jointly optimizing for the text embeddings~\cite{chung2024prompttuning}, and one can use the text condition as the control knob for mode selection~\cite{kim2023regularization,kim2025dreamsampler}. However, the existing methods are very slow to execute~\cite{chung2024prompttuning}, require heuristic tuning of the algorithm that is targetted for specific appcliations~\cite{kim2023regularization,kim2025dreamsampler}, and often yields over-emphasized results that render the results closer to image editing. Moreover, the situation is inherently different for image restoration and medical image reconstruction.
In most image restoration tasks, meaningful text conditioning is scarce because such tasks often lack associated textual metadata that describe the imaging process or the specific content being restored, necessitating algorithms like prompt-tuning to automatically generate useful prompts~\cite{chung2024prompttuning}.

In contrast to low-level vision problems explored in~\cite{kim2023regularization,chung2024prompttuning,kim2025dreamsampler}, in medical imaging, ample metadata exists in various forms. Patient demographics, medical history, body part, imaging parameters of the scan, etc., all provide valuable additional information, which can reshape the prior by conditioning $p(\x|\cb)$. These factors have been surprisingly overlooked in the context of inverse problems, where only the measurement signal is used to retrieve the clean signal, a suboptimal practice according to the data processing inequality~\cite{cover1999elements}.

Most prior works leveraging text-conditioned generative models for medical data rely on relatively simple and coarse-grained conditioning that is interpretable and often acts closer to discrete class conditioning.
Roentgen~\cite{bluethgen2024vision} finetunes Stable Diffusion for X-ray synthesis from simple text descriptions of the pathology. Kim {\em et al.}~\cite{kim2024controllable} trains a text-to-image LDM for BRATS~\cite{menze2014multimodal} brain MRI conditioned on contrast and some pathology information. Medisyn~\cite{cho2024medisyn} trains W{\"u}rstchen~\cite{pernias2023wurstchen}, a multi-scale text-to-video LDM on a highly diverse set of data including X-ray, CT, MRI, endoscopy, OCT, etc., but on relatively simpler prompts focused on specifying the modality of the target image. Pinaya {\em et al.}~\cite{pinaya2022brain} trains a 3D LDM that is conditioned on numerical values of age, sex, and ventricular size to generate T1w brain images (UK Biobank~\cite{sudlow2015uk}). GenerateCT~\cite{hamamci2023generatect} trains a two-stage discrete-continuous text-to-volume diffusion model, which takes in as input age, sex, and impression (i.e. pathology) in text form.
The work that is possibly the closest to our work is TUMSyn~\cite{wang2024towards}, which is a conditional neural implicit model trained specifically for MRI contrast conversion. The text condition of TUMSyn includes fine-grained information including patient information, scanner type, MR imaging parameters, etc. 
However, TUMSyn is a model tailored for a specific task of contrast conversion. In contrast, our work focuses on first constructing a general-purpose diffusion generative prior, then using this prior in a zero-shot fashion for solving inverse problems. While we focus on compressed sensing MRI (CS-MRI), our method can also be used for unconditional data synthesis, image inpainting, etc.
Another important point to note is that even for text conditional generative models that were trained specifically for the MRI modality~\cite{kim2024controllable,pinaya2022brain,wang2024towards}, {\em none} of the works considered the complex-valued nature of MR images, and hence cannot be used for native tasks such as CS-MRI. Moreover, the use of these generative models was mostly constrained to data augmentation~\cite{kim2024controllable,pinaya2022brain}, or the task that the model was trained for~\cite{wang2024towards}. 
In what follows, we demonstrate that a pre-trained generative model, which was not specifically trained for any particular MRI inverse problem, can be adapted at inference time to various DIS solvers. This approach achieves a reconstruction quality that surpasses what was previously attainable with unconditional models.

To leverage text conditioning for inverse problems in medical imaging, we propose using a pre-trained generative model that is conditioned on text. 
Several recent works~\cite{bluethgen2024vision,khader2023denoising,pinaya2022brain,kim2024controllable,cho2024medisyn} focus on developing text-to-image diffusion models for image synthesis. However, most of these works focus on improving the generation quality of the images, and the use of the generated data is limited to data augmentation. For the methods that target MRI specifically~\cite{pinaya2022brain,kim2024controllable}, the models are conditioned by simple text prompts similar to the role of classifiers.
A concurrent work of TumSyn~\cite{wang2024towards} trains an implicit neural network for MRI contrast conversion on detailed metadata, but is constrained to a specific task and does not handle the complex-valued nature of MR imaging.

In this work, we propose \method, the first text-to-image MRI diffusion model that is conditioned on detailed metadata including patient demographics (e.g. age, sex), MR imaging parameters (e.g. TR, TE, etc.), anatomy (e.g. knee, brain), contrast (e.g. PDFS, T2FLAIR), slice location, and pathological findings. Our model is trained on minimum variance unbiased estimate (MVUE)~\cite{jalal2021robust} complex-valued images derived from raw $k$-space data and estimated sensitivity maps~\cite{uecker2014espirit} so that it can be naturally used for downstream tasks such as CS-MRI. Using \method~as a prior, we propose a new paradigm of metadata-conditioned CS-MRI, showing that conditioning the inverse problem-solving process with metadata significantly improves the reconstruction performance, gradually improving with more information.
Importantly, we show that \method~consistently outperforms the unconditional counterparts regardless of the circumstances. 

\section{Result}
\label{sec:result}

\subsection{\method~captures intricate metadata cues}
\label{sec:result_image_synthesis}

A vital requirement for text-conditioned diffusion models in inverse problems is their ability to generate realistic images aligned with the given metadata. Figure~\ref{fig:image_synthesis} demonstrates this alignment qualitatively. We extract metadata from real MRI volumes and feed it into our diffusion model at a CFG scale of 5\footnote{CFG scale of 1 is naive conditional inference. Increasing this value emphasizes the conditioning.} to emphasize the conditioning signal. Note that no measurement signal from the original image was used in the generation of the images in the second row of Figure~\ref{fig:image_synthesis}.
Notably, the synthesized images exhibit anatomically coherent structures and contrast weightings consistent with the metadata prompts. For instance, T1POST brain slices display dark cerebrospinal fluid and a bright dural rim, whereas T2FLAIR-like slices depict dark ventricles and sulci against relatively brighter parenchyma. In proton-density (PD) knee scans, bright fat signals, and sharp tissue delineation emerge, while PD fat-suppressed (PDFS) acquisitions suppress fat but preserve fluid hyperintensities. These examples confirm that \method~correctly interprets slice-level metadata, producing anatomically plausible images reflecting the instructed contrast and tissue characteristics.

\begin{table}[!t]
    \centering
    \renewcommand{\arraystretch}{1.2} 
    \setlength{\tabcolsep}{4pt}      
    \resizebox{\textwidth}{!}{%
    \begin{tabular}{@{}ccccccccccccc@{}}  
    \toprule
    \rowcolor{lightc0}
    & \multicolumn{12}{c}{\textbf{Knee}} \\
    \midrule
    \rowcolor{lightc0}
    \textbf{Mask type} & \multicolumn{6}{c}{\textbf{Uniform 1D}} & \multicolumn{6}{c}{\textbf{Poisson 2D}} \\
    \cmidrule(lr){2-7} \cmidrule(lr){8-13}
    \rowcolor{lightc1}
    \textbf{Acc. factor} & \multicolumn{3}{c}{$\times 4$} & \multicolumn{3}{c}{$\times 8$} 
                         & \multicolumn{3}{c}{$\times 8$} & \multicolumn{3}{c}{$\times 15$} \\
    \cmidrule(lr){2-4} \cmidrule(lr){5-7} \cmidrule(lr){8-10} \cmidrule(lr){11-13}
    \rowcolor{lightc1}
    \textbf{Metric} & \textbf{PSNR$\uparrow$} & \textbf{SSIM$\cdot10^2\uparrow$} & \textbf{LPIPS$\cdot10^2\downarrow$}
                    & \textbf{PSNR$\uparrow$} & \textbf{SSIM$\cdot10^2\uparrow$} & \textbf{LPIPS$\cdot10^2\downarrow$}
                    & \textbf{PSNR$\uparrow$} & \textbf{SSIM$\cdot10^2\uparrow$} & \textbf{LPIPS$\cdot10^2\downarrow$}
                    & \textbf{PSNR$\uparrow$} & \textbf{SSIM$\cdot10^2\uparrow$} & \textbf{LPIPS$\cdot10^2\downarrow$} \\

    \midrule
    0.0 (Uncond) 
    & 30.52 \stddev{2.84} & 87.02 \stddev{8.83} & 20.97 \stddev{4.64}
    & 28.15 \stddev{2.27} & 81.97 \stddev{7.33} & 25.94 \stddev{3.05}
    & 34.45 \stddev{4.65} & 91.50 \stddev{12.1} & 12.46 \stddev{8.57}
    & 32.61 \stddev{3.31} & 89.66 \stddev{7.80} & 15.82 \stddev{3.69}\\

    1.0 
    & 30.66 \stddev{2.88} & 87.35 \stddev{9.15} & 20.58 \stddev{4.67}
    & 28.32 \stddev{2.30} & 82.71 \stddev{7.39} & 25.39 \stddev{3.30}
    & \textbf{34.55} \stddev{4.83} & \textbf{91.51} \stddev{12.8} & \textbf{12.40} \stddev{8.66}
    & 32.64 \stddev{3.56} & 89.80 \stddev{9.39} & 15.75 \stddev{5.36}\\

    2.0 
    & 30.83 \stddev{2.78} & 87.96 \stddev{8.27} & 20.11 \stddev{4.20}
    & 28.45 \stddev{2.25} & 83.32 \stddev{6.97} & 25.00 \stddev{3.12}
    & 34.43 \stddev{5.17} & 90.88 \stddev{14.6} & 12.92 \stddev{10.1}
    & 32.73 \stddev{3.53} & 89.90 \stddev{9.00} & 15.67 \stddev{5.03}\\

    3.0 
    & \textbf{30.92} \stddev{2.64} & \textbf{88.30} \stddev{7.15} & \textbf{19.91} \stddev{3.56}
    & \textbf{28.47} \stddev{2.30} & \textbf{83.51} \stddev{7.31} & \textbf{24.99} \stddev{3.49}
    & 33.98 \stddev{5.85} & 89.51 \stddev{17.1} & 14.22 \stddev{12.5}
    & \textbf{32.75} \stddev{3.57} & \textbf{89.94} \stddev{9.19} & \textbf{15.66} \stddev{5.25}\\
    \midrule
    \rowcolor{lightc0}
    & \multicolumn{12}{c}{\textbf{Brain}} \\
    \midrule
    \rowcolor{lightc0}
    \textbf{Mask type} & \multicolumn{6}{c}{\textbf{Uniform 1D}} & \multicolumn{6}{c}{\textbf{Poisson 2D}} \\
    \cmidrule(lr){2-7} \cmidrule(lr){8-13}
    \rowcolor{lightc1}
    \textbf{Acc. factor} & \multicolumn{3}{c}{$\times 4$} & \multicolumn{3}{c}{$\times 8$}
                         & \multicolumn{3}{c}{$\times 8$} & \multicolumn{3}{c}{$\times 15$} \\
    \cmidrule(lr){2-4} \cmidrule(lr){5-7} \cmidrule(lr){8-10} \cmidrule(lr){11-13}
    \rowcolor{lightc1}
    \textbf{Metric} & \textbf{PSNR$\uparrow$} & \textbf{SSIM$\cdot10^2\uparrow$} & \textbf{LPIPS$\cdot10^2\downarrow$}
                    & \textbf{PSNR$\uparrow$} & \textbf{SSIM$\cdot10^2\uparrow$} & \textbf{LPIPS$\cdot10^2\downarrow$}
                    & \textbf{PSNR$\uparrow$} & \textbf{SSIM$\cdot10^2\uparrow$} & \textbf{LPIPS$\cdot10^2\downarrow$}
                    & \textbf{PSNR$\uparrow$} & \textbf{SSIM$\cdot10^2\uparrow$} & \textbf{LPIPS$\cdot10^2\downarrow$} \\
    \midrule
    0.0 (Uncond)
    & 32.15 \stddev{3.78} & 90.37 \stddev{10.1} & 16.04 \stddev{5.23}
    & 29.59 \stddev{3.23} & 86.75 \stddev{9.07} & 19.77 \stddev{3.86}
    & 36.51 \stddev{5.24} & 93.30 \stddev{13.6} & 11.08 \stddev{8.42}
    & 34.62 \stddev{4.06} & 92.47 \stddev{10.2} & 12.82 \stddev{5.00}\\

    1.0 
    & 32.57 \stddev{3.32} & 90.85 \stddev{6.78} & \textbf{15.20} \stddev{3.10}
    & 29.71 \stddev{3.20} & 87.47 \stddev{8.78} & \textbf{19.48} \stddev{3.79}
    & \textbf{36.75} \stddev{5.18} & \textbf{93.75} \stddev{13.0} & \textbf{10.83} \stddev{8.31}
    & 34.78 \stddev{4.10} & 92.75 \stddev{10.6} & 12.71 \stddev{5.40}\\

    2.0 
    & \textbf{32.58} \stddev{3.60} & 91.53 \stddev{9.24} & 15.34 \stddev{4.70}
    & \textbf{29.75} \stddev{3.27} & \textbf{87.60} \stddev{9.20} & 19.72 \stddev{4.30}
    & 36.72 \stddev{5.31} & 93.49 \stddev{13.7} & 10.99 \stddev{8.64}
    & \textbf{34.87} \stddev{4.06} & \textbf{92.95} \stddev{10.4} & \textbf{12.61} \stddev{5.21}\\

    3.0 
    & 32.47 \stddev{3.35} & \textbf{91.76} \stddev{7.09} & 15.63 \stddev{3.65}
    & 29.66 \stddev{3.42} & 86.76 \stddev{9.51} & 20.97 \stddev{4.96}
    & 36.37 \stddev{5.87} & 92.27 \stddev{16.0} & 11.90 \stddev{10.7}
    & 34.70 \stddev{4.52} & 92.26 \stddev{12.7} & 13.24 \stddev{7.76}\\

    \bottomrule
    \end{tabular}
    }
    \caption{\bf\footnotesize Quantitative results ($\mu \pm \sigma$) for the fastMRI datasets with different masking patterns (Uniform 1D, Poisson 2D) and acceleration factors ($\times 4$, $\times 8$) by varying the conditioning strength (CFG).}
    \label{tab:main_results}
\end{table}

\subsection{Conditional inference outperforms unconditional inference}
\label{sec:result_conditional_inference}

We first focus on the case where we have access to all the metadata that was used during training, including anatomy, slice, contrast, and MR imaging parameters.
Table~\ref{tab:main_results} reports the quantitative performance of \method~for CS-MRI reconstruction under various acceleration factors, undersampling masks, and anatomies, while Figure\ref{fig:main_results} provides a qualitative summary.
We test different CFG scales, from 0.0 (unconditional) to 3.0, and generally see better reconstruction accuracy at higher scales, although in some runs the best performance occurs at lower guidance strengths.
We note that when we increase the CFG scale beyond 3.0, the performance starts to drop, which is expected from the known limitations of using high guidance scales~\cite{poole2022dreamfusion,bradley2024classifier}, especially when coupled with inverse problem solving~\cite{chung2024cfg++}. This indicates that metadata injection augments the model's prior beyond information already contained in the undersampled measurement. Furthermore, pushing CFG to around 2--3 provides the best trade-off between leveraging metadata and avoiding over-reliance on text prompts.
When pitted against another diffusion model of similar parameter count\footnote{Both are U-Net architecture with about 400M parameters.} used in the original Decomposed Diffusion Sampler (DDS) framework~\cite{chung2024decomposed}, \method~achieves comparable performance in the unconditional setting. However, once metadata conditioning is enabled, \method~significantly surpasses the DDS baseline. This trend remains consistent regardless of anatomy, mask type, or sampling ratio. Moreover, conditional inference often reduces variability across slices. For example, the structural similarity index measure (SSIM)~\cite{Wang2004} and the learned perceptual image patch similarity (LPIPS)~\cite{zhang2018unreasonable} scores under conditional settings tend to have similar or lower standard deviations compared to the unconditional baseline. This implies that metadata conditioning not only improves average performance but also leads to more stable and consistent reconstructions from slice to slice.

\subsection{Impact of MR Imaging Parameters}

\begin{wrapfigure}[18]{r}{0.6\textwidth}
    \centering
    \vspace{-0.5cm}
    \includegraphics[width=\linewidth]{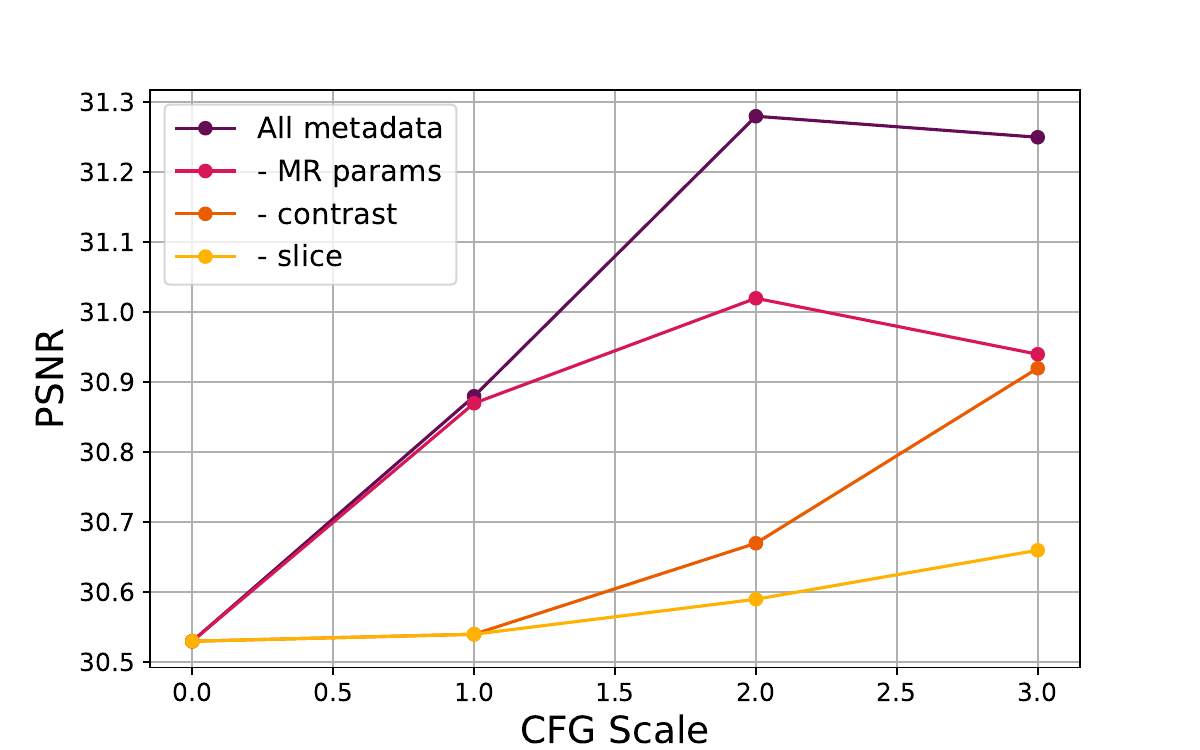}
    \caption{\bf\footnotesize PSNR vs. CFG by varying the amount of information contained in the metadata. Experiments are conducted on a subset of fastMRI knee data with uniform 1D $\times 4$ acceleration.}
    \label{fig:varying_information}
\end{wrapfigure}

It may be clear from the image synthesis experiment that the obviously human-interpretable metadata including the anatomy, slice location, and contrast will have a significant impact on the image reconstruction procedure.
In contrast, the difference that arises from changing the MR imaging parameters such as TR, TE, TI, and FA are rather subtle and hard to interpret directly.
However, including MR imaging parameters such as TR, TE, TI, and flip angle provides direct insight into the specific physical conditions under which the MRI signal was generated. Each parameter influences how the magnetization of tissues recovers and decays over time, shaping tissue contrasts and intensity patterns in predictable ways. For instance, shorter TRs may highlight certain tissues at the expense of others, while longer TEs can amplify T2-weighted contrasts. 
TI and flip angle further adjust the equilibrium states of magnetization and influence how spins align with or deviate from the main magnetic field. 

\begin{wrapfigure}[17]{r}{0.6\textwidth}
    \centering
    \vspace{-0.5cm}
    \includegraphics[width=\linewidth]{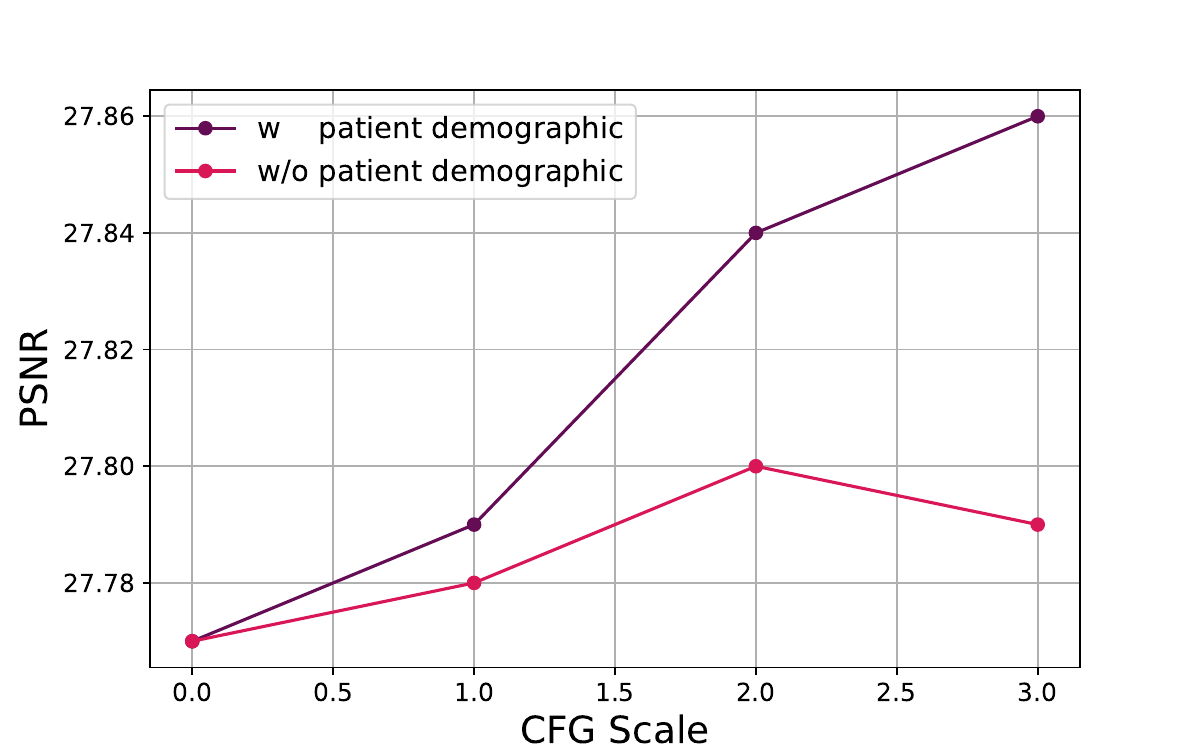}
    \caption{\bf\footnotesize PSNR vs. CFG on SKM-TEA validation set with and without patient demographic (i.e. age, sex) information.}
    \label{fig:skm_tea}
\end{wrapfigure}

In essence, these parameters represent the ``recipe'' of the pulse sequence that determines how different tissue components appear in the final image. By encoding these parameters as part of the prior, a model can better anticipate the expected distribution of intensity values and subtle contrasts that arise from particular pulse sequences. Consequently, this additional context helps the model resolve ambiguities from undersampled data more effectively, guiding the reconstruction process toward solutions that are consistent with the known physics of MR image formation.
In Fig.~\ref{fig:varying_information}, we show that there is a significant gap between the case with and without this condition. 
Here, ``All metadata'' corresponds to the case where we use all the ground truth metadata, as presented in the previous ``Conditional inference outperforms unconditional inference'' section; ``- MR params'' refers to the case where TR, TE, TI, and FA information were removed; ``- contrast'', ``- slice'' are the cases where the contrast and slice information were further removed, respectively.
As we remove more and more information from the input text prompt, we see a gradual decrease in the reconstruction performance, highlighting the advantage of \method~arising from the data processing inequality.

\subsection{Patient Demographics as a Reconstruction Prior}

Age-dependent tissue heterogeneity, variations in organ morphology, and progressive degenerative changes all directly influence underlying MR signal characteristics. Likewise, sex-based physiological distinctions—like hormonal fluctuations and structural differences in muscle, bone, and fat distribution—are known to modulate intensity and contrast in MR images. 
By injecting this demographic context, the diffusion model learns more nuanced priors about the expected image anatomy and contrast patterns. To validate this hypothesis, we fine-tuned our model on SKM-TEA~\cite{desai2022skm} dataset which also contains patient demographic as the metadata, and tested the performance of CS-MRI on the SKM-TEA validation set with uniform1D $\times 8$ downsampling. In Fig.~\ref{fig:skm_tea}, we see the reconstruction performance in Peak signal-to-noise-ratio (PSNR) with and without the patient demographic, validating our hypothesis.

\subsection{Pathology Priors and Robustness to Misinformation}

\begin{figure}[!t]
    \centering
    \makebox[\linewidth][c]{%
        \includegraphics[width=1.2\linewidth]{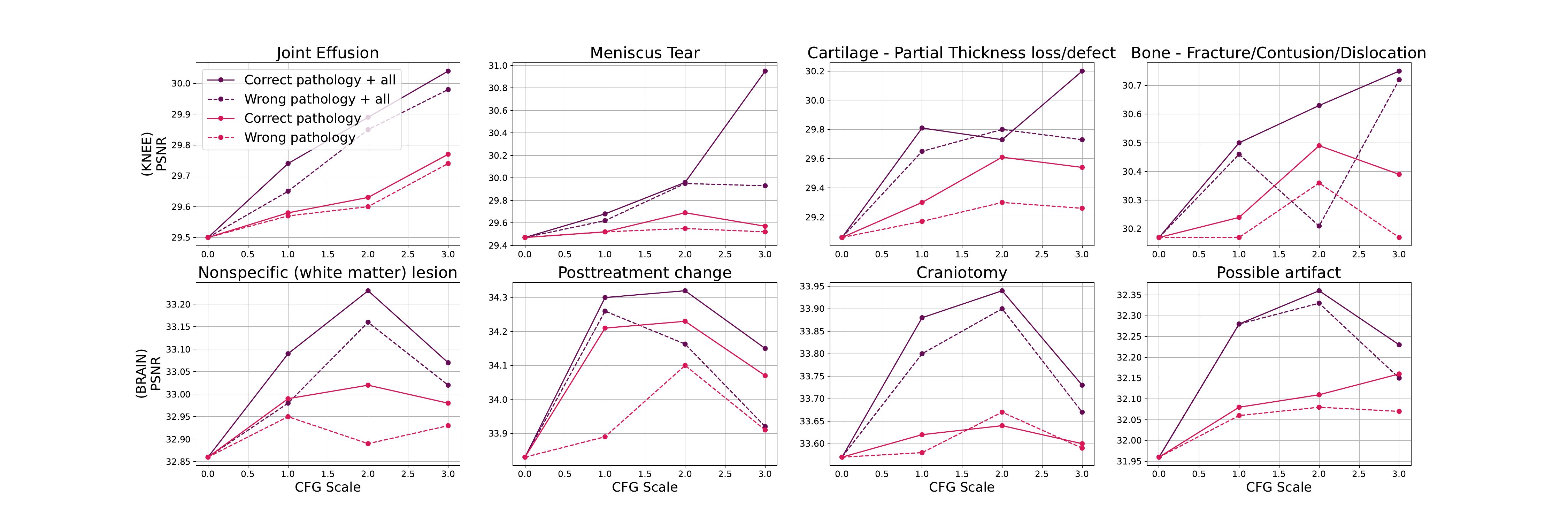}
    }
    \caption{\bf\footnotesize Trend in PSNR vs. CFG when the pathology of the scan is known. The first row shows 4 different pathologies that are the most dominant in the fastMRI knee dataset. The second row shows 4 different pathologies in the fastMRI brain dataset. All experiments are done with uniform 1D $\times 4$ acceleration.}
    \label{fig:pathology_plot}
\end{figure}

In clinical workflow, there may be cases where the impressions of the pathological finding can be predicted a priori to the MR scan. For instance, a patient may have gone through a previous X-ray or CT scan, where a classic bone lesion was evident on the scan so that MRI ends up being a formality to nail down the specifics rather than to discover an entirely new diagnosis. Moreover, if the patient's symptoms, physical exam, and labs scream a particular diagnosis, the radiologist can already guess the underlying pathological findings of the MRI scan. To verify the effectiveness of \method~in such a case, we tested two different situations: when we only use the information about the pathological finding in the slice (Correct pathology), and using the pathological finding information along with all the other metadata (Correct pathology + all). In Fig.~\ref{fig:pathology_plot}, we show the trend in PSNR vs. CFG guidance scale on 4 different pathological finding types that were the most dominant in the fastMRI+ dataset for each anatomy. We note that for the less dominant pathologies, the size of the dataset is too small (in the order of hundreds) for the model to properly learn the characteristic, so we exclude the other pathological findings from our experiment. In the figure, we show that having access to the pathological finding information helps both when we only use the pathological finding information and in the case where we have all the metadata along with it.

One concern with this approach could be when the attending clinician mistakenly predicts a wrong pathological finding impression - will this hamper the reconstruction quality by hallucinating a non-existent lesion? To test the robustness of our approach in such a case, in Fig.~\ref{fig:pathology_plot}, we also show with the dotted line, the case where we randomly swapped the ground truth pathology with a different one. From the plot, we see that while using the wrong pathological finding information has inferior performance than the case where we have the correct information, the results are still better than the unconditional case. 
From this experiment, we see that even when the prediction of the exact pathological finding is wrong, knowing the initial impression of the attending physician can still be helpful in the reconstruction.
We conjecture that even an incorrect pathology label can still help the model distinguish a “healthy” subject from one harboring “some” abnormality. By recognizing the presence (rather than the absence) of a lesion, the model learns to focus its reconstruction on suspicious areas. This partial guidance, though misaligned in nature, still provides a net boost over the unconditional baseline.

\section{Discussion}
\label{sec:discussion}

There has been a recent surge of interest in developing text-to-image diffusion models for medical images, which can potentially be used for diverse clinical practice applications. However, much of the discourse has revolved around synthetic data generation as a primary use case. Synthetic data is often positioned as a generative augmentation tool to improve the performance of downstream tasks, such as segmentation and classification, by enhancing data diversity~\cite{bluethgen2024vision,hamamci2023generatect}.

Beyond augmentation, the full potential of pre-trained text-to-image diffusion models in medical imaging remains underexplored. In computer vision, these models have demonstrated remarkable versatility, being successfully adapted for tasks like data mining~\cite{siglidis2024diffusion}, 3D scene reconstruction~\cite{chung2023luciddreamer}, and solving inverse problems~\cite{chung2023diffusion}, in a zero-shot fashion. The extension of these methodologies to medical imaging could similarly revolutionize clinical workflows. For instance, pre-trained text-to-image models could be re-purposed to generate anatomically consistent counterfactuals for diagnostic reasoning, guide 3D reconstruction of volumetric scans, or act as priors in solving inverse problems such as accelerated MRI reconstruction. Each of these applications could address practical limitations in current clinical practice, paving the way for more integrated, intelligent systems in healthcare.

In this work, we tackle a specific inverse problem, CS-MRI, as we believe it is the most imminent application that can benefit from text conditioning with metadata. Extensive experiments show evidence that metadata that is often ignored in clinical practice can indeed be useful. While we believe that this is an important first step, there are several limitations.

First, note that CLIP~\cite{radford2021learning}, the text encoder used for \method, was used without fine-tuning it to MRI image-text pairs. Naturally, it may not {\em understand} the texts, potentially making it harder for the diffusion model to learn the relation between the given metadata and images. An interesting future study will involve designing and training a text encoder specific to MRI or incorporating large language models (LLM) that were pre-trained on medical data as a text encoder.

Second, a key limitation of leveraging text-to-image diffusion models for CS-MRI lies in the heterogeneity of metadata across datasets and clinical settings. Available metadata often varies depending on the source of the data. For instance, datasets such as SKM-TEA~\cite{desai2022skm} provide patient-specific details like age and sex, whereas fastMRI~\cite{zbontar2018fastmri} primarily includes imaging parameters and lacks patient demographic information. This variability complicates the design of a unified structural format for input metadata, which is crucial for effectively conditioning the generative model. 

To address this challenge, future research should focus on developing robust frameworks for handling incomplete or inconsistent metadata. For instance, developing flexible metadata encoding schemes to process varying types and quantities of metadata would be an interesting direction. In order to scale this to clinical workflow, it would also be crucial to standardize the metadata schema, which would involve collaboration between researchers, clinicians, and database curators.

\section{Method}

\subsection{Data curation}

Our main dataset collection is based on the fastMRI~\cite{zbontar2018fastmri} data on the knee and brain. From the raw $k$-space data of varying size, we compute the inverse Fourier transform, and center-crop all images to resolution $320 \times 320$, so that the unwanted boundaries are eliminated. We take the Fourier transform back to the $k$-space and estimate the sensitivity maps through ESPiRiT~\cite{uecker2014espirit} to compute the MVUE estimate. We normalize the images by dividing the complex image with 99\% quantile, which approximately provides the range of $[-1.5, 1.5]$ for both the real and the imaginary channels. Our training set consists of 167,375 slices across 8,344 volumes.

For the patient demographic experiment, we fine-tune our model using the SKM-TEA~\cite{desai2022skm} dataset, where the images are of $512 \times 512$ resolution. We use the same data preprocessing steps as in the fastMRI case and directly use the sensitivity maps provided in the dataset. The training consists of 25,731 slices of images across 139 volumes.

\begin{table}[!thb]
\centering
\caption{Parameter ranges/values for each dataset}
\label{tab:dataset-params}
\resizebox{0.8\textwidth}{!}{%
\begin{tabular}{l|l|l|l}
\hline
\textbf{Parameter} & \textbf{fastMRI Knee} & \textbf{fastMRI Brain} & \textbf{SKM-TEA} \\ 
\hline
Anatomy & Knee & Brain & Knee \\
Slice index & 0--30 & 0--25 & 0--159 \\
Contrast & PD, PDFS & T1, T1PRE, T1POST, T2, FLAIR & PD, T2 \\
Pathology & 
See Supp. Fig.~\ref{fig:metadata_statistic_knee} & See Supp. Fig.~\ref{fig:metadata_statistic_brain} & See Supp. Fig.~\ref{fig:metadata_skm_tea} \\
Sequence & Turbospinecho & Turbospinecho, Flash & QDess \\
TR (ms) & 2000--3930 & 247--15810 & 18.176--20.36 \\
TE (ms) & 24--35 & 2--126 & 5.796--6.428 \\
TI (ms) & 100 & 100--2500 & - \\
Flip angle ($^\circ$) & 122--150 & 69--180 & 20 \\
\hline
\end{tabular}
}
\end{table}

\subsection{Metadata conditioning}

For the metadata, we consider the following entries: \code{anatomy}, \code{slice}, \code{contrast}, \code{pathology}, \code{sequence}, \code{TR}, \code{TE}, \code{TI}, \code{flip angle}. These are structured into a string, similar to the following example: ``\code{Knee, Slice 19, PDFS, Pathology: Displaced Meniscal Tissue, Meniscus Tear, Bone-Subchondral edema, TR: 3150, TE: 33, TI: 100, Flip angle: 150}''. All the metadata except for the pathology information is extracted from the metadata provided directly in the fastMRI database. See Tab.~\ref{tab:dataset-params}, Fig.~\ref{fig:metadata_statistic_knee}, Fig.~\ref{fig:metadata_statistic_brain}, and Fig.~\ref{fig:metadata_skm_tea} for the metadata conditions used, and their statistics. In the fastMRI+~\cite{zhao2021fastmri+} database, there are annotations with bounding boxes whenever a pathology can be seen in the specified slice. We ignore the location of the bounding box and only append the existing types of pathologies that can be found in the slice, separated with a comma. When no pathologies are found, we do not append this trait. To enhance the robustness for the cases where some of the metadata might not be provided during inference, we randomly drop the MR imaging parameters TR, TE, TI, and Flip angle with 50\% probability during training.

The metadata for SKM-TEA was structured with the following entries: \code{sequence}, \code{anatomy}, \code{slice}, \code{age}, \code{sex}, \code{pathology}, \code{sequence type}, \code{TR}, \code{TE}, and \code{flip angle}. This is an example of the structured text based on the described metadata: ``\code{Qdess, Knee, Slice 63, Age: 61, Sex: M, Pathology: Cartilage Lesion, TR: 18.352, TE: 5.876, Flip Angle: 20}''. To ensure the model can easily disregard previously introduced text condition information, the sequence was prioritized and placed at the beginning of the metadata. Similar to fastMRI, bounding box information was omitted, and the same methodology was applied. For pathology, the original annotations provided 3D bounding boxes for volumes. These were decomposed into corresponding 2D bounding boxes along the z-axis for each slice, and the relevant slices were included. Additionally, only pathology information with a confidence score of 4 or higher on a 0-5 scale was retained.

\subsection{Diffusion model training}

Consider a continuous sequence of densities $p_t(\x)$, where $t$ indicates time in a diffusion process. Let $\x_0 \sim p_0(\x)$ be the data distribution, and $\x_T = \Nc(0, I_d)$ be the reference Gaussian normal distribution. In diffusion models~\cite{ho2020denoising,song2020score}, the data distribution is gradually corrupted with Gaussian noise with the perturbation kernel $p(\x_t|\x_0) = \Nc(\x_0; 0, t^2I_d)$, until it approximates a standard normal distribution.
Diffusion models learn the denoising generative reverse process, i.e. creating data from noise. This process is governed by the score function $\nabla_{\x_t} \log p(\x_t)$~\cite{hyvarinen2005estimation}, which is estimated through denoising score matching $\s_\theta(\x_t) \approx \nabla_{\x_t} \log p(\x_t)$~\cite{hyvarinen2005estimation}.
Text conditional diffusion models~\cite{rombach2022high,deepfloyd_if} are trained with image-text pairs with random dropping so that it is possible to both sample from $p(\x)$ by using $\epsilonb_\theta(\x_t)$ or from $p(\x|\cb)$ by using $\epsilonb_\theta(\x_t, \cb)$, where $\cb$ is the text embedding. Further, to emphasize the conditioning signal, another widely used technique is classifier free guidance (CFG)~\cite{ho2021classifierfree}, where $\epsilonb_\theta^\gamma(\x_t, \cb) := \epsilonb_\theta(\x_t) + \gamma(\epsilonb_\theta(\x_t, \cb) - \epsilonb_\theta(\x_t))$ is used to sample from the sharpened posterior $p(\x)p(\cb|\x)^\gamma$.

As the loss function, we use the epsilon-matching loss function commonly used in DDPMs~\cite{ho2020denoising} and employ a fixed variance setting rather than learning the variance. The model parameters are optimized for 50 epochs using the AdamW optimizer with a learning rate of 0.0001. All experiments were conducted using PyTorch 2.1 in Python with CUDA 12.1, running on 8 NVIDIA H100 GPUs (80GB). The effective batch size was set to 64.

We fine-tuned the above pixel-level diffusion model on the SKM-TEA dataset with a modified input resolution of 512 × 512, as opposed to the original 320 × 320 size. To accommodate the increased computational demand, we adjusted the effective batch size to 16. All other training settings remained the same. Fine-tuning was conducted for an additional 30 epochs using PyTorch 2.1 with CUDA 12.1 on 8 NVIDIA H100 GPUs (80GB).

\subsection{Diffusion model architecture}

We utilize a pixel-level diffusion model for both unconditional and metadata-conditional generation. The decision to employ pixel-level diffusion, rather than latent-space diffusion via a Variational Autoencoder (VAE)~\cite{kingma2013auto}, stems from the inherent limitations of VAEs. Specifically, VAE-based reconstruction often compromises high-frequency details~\cite{elata2024novel}, and solving inverse problems in the latent space requires decoding to ensure data consistency. While such methods~\cite{chung2024prompttuning, rout2024solving, kim2023regularization} are theoretically feasible, they frequently fail in practice and lack consensus as a reliable solution. Consequently, we opted to perform generation directly in pixel space.

To train the diffusion model on MRI complex values, we separate the real and imaginary parts, concatenating them along the channel dimension to create a 2-channel input. This results in a 2D input of size 2 × 320 × 320, which is used during training.
Our network is adapted from the DeepFloyd IF~\cite{deepfloyd_if} pixel-level text-to-image diffusion model, with modifications to accommodate our input format by adjusting the number of channels in the layers. Unlike natural image generation, the textual metadata associated with MRI data is significantly smaller. Therefore, we use the CLIP text encoder instead of the T5-XXL encoder employed in DeepFloyd IF, offering a more compact and memory-efficient representation of the metadata. During training, the CLIP text encoder is frozen, and only the parameters of the UNet modules are learnable.

\subsection{Inverse problem solving with diffusion models}

We are interested in linear inverse problems of the following form
\begin{align}
\label{eq:ip}
    \y = A\x + \n, \quad \n \sim \Nc(0, \sigma_y^2 I_d),\, \x \in \Cd^d, \y \in \Cd^n,
\end{align}
where for the multi-coil MRI case~\cite{deshmane2012parallel}, the forward model is defined in parallel for each measurement coil
\begin{align}
    \y_i = DFS_i\x + \n,
\end{align}
where $D$ is the downsampling operator, $F$ is the 2D discrete Fourier transform matrix, and $S_i$ is the sensitivity map. Posterior sampling approaches with diffusion models often utilize Bayes rule
\begin{align}
\label{eq:bayes}
    \nabla_{\x_t} \log p(\x_t|\y) = \nabla_{\x_t} \log p(\x_t) + \nabla_{\x_t} \log p(\y|\x_t),
\end{align}
to write the score function of the posterior in terms of the prior score function. 
A vast literature on DIS can be structured as different ways of approximating the intractable time-dependent likelihood $\nabla_{\x_t} \log p(\y|\x_t)$~\cite{daras2024survey}.
A few recent works~\cite{chung2024prompttuning,kim2023regularization,kim2025dreamsampler} tried to incorporate text conditions through Stable Diffusion~\cite{rombach2022high}. Among these, P2L~\cite{chung2024prompttuning} is prohibitively slow and is highly unstable when used with CFG. TReg~\cite{kim2023regularization} is tailored for severe degradations and produces cartoon-like artifacts. Dreamsampler~\cite{kim2025dreamsampler} suggests tailored algorithms for specific types of inverse problems. Generally, setting moderately high values of CFG leads to a steep decrease in the stability of all the algorithms. None of the existing methods have shown the stability and performance required to extend the method to the medical imaging setting, where this is crucial.

We opt for a simple, established baseline for our choice of diffusion model-based inverse problem solver. For this, we choose DDS~\cite{chung2024decomposed}, which is known to be especially robust to large-scale medical imaging inverse problems. Specifically, with a variance preserving form, DDS with DDIM sampling~\cite{song2020denoising} can be described as iterating the following steps for $t = T, T-1, \dots, 1$ initialized with random Gaussian noise.
\begin{align}
    \hat\x_{0|t} &= \frac{1}{\sqrt{\bar\alpha_t}} \left(
    \x_t - \sqrt{1 - \bar\alpha_t}\epsilonb_\theta^\gamma(\x_t, \cb)
    \right) \\
    \hat\x'_{0|t} &= \argmin_{\x} \frac{\xi}{2}\|\y - A\x\|_2^2 + \frac{1}{2}\|\x - \hat\x_{0|t}\|_2^2 \label{eq:prox_opt} \\
    \x_{t-1} &= \sqrt{\bar\alpha_{t-1}}\hat\x'_{0|t} + \sqrt{1 - \bar\alpha_{t-1} - \sigma_t^2}\epsilonb_\theta^\gamma(\x_t, \cb) + \sigma_t\epsilonb_t,\, \epsilonb_t \sim \Nc(0, I_d)
\end{align}
where $\sigma_t := \eta\sqrt{(1 - \alpha_t)(1 - \bar\alpha_{t-1})/(1 - \bar\alpha_t)}$ with a constant hyperparameter $\eta = 0.8$. 
From a high-level viewpoint, skipping the data consistency optimization step in \eqref{eq:prox_opt} would lead to DDIM sampling from the prior distribution $p(\x)$. In order to sample from the approximate posterior, DDS modulates the posterior mean $\hat\x_{0|t} = \Ed[\x_0|\x_t]$ with data consistency steps to approximate the conditional posterior mean $\hat\x'_{0|t} \approx \Ed[\x_0|\x_t, \y]$.
Specifically, for the optimization problem in \eqref{eq:prox_opt}, the key of DDS is to use few-step conjugate gradient steps. In our case, we use 5-step update with $\xi = 5.0$.

Our method is tested on two different undersampling patterns - uniform1D as implemented in fastMRI~\cite{zbontar2018fastmri}, and poisson2D as implemented in sigpy~\cite{ong2019sigpy}. For uniform1D, we test two different acceleration factors: $\times 4$ with 8\% autocalibrating signal (ACS) region in the center, and $\times 8$ with 4\% ACS region. For poisson2D, we test $\times 8$ and $\times 15$ acceleration.
For quantitative evaluation, we use three standard metrics - PSNR, SSIM, and LPIPS.

\printbibliography

\pagebreak
\appendix

\begin{figure}[!h]
    \centering
    \includegraphics[width=\linewidth]{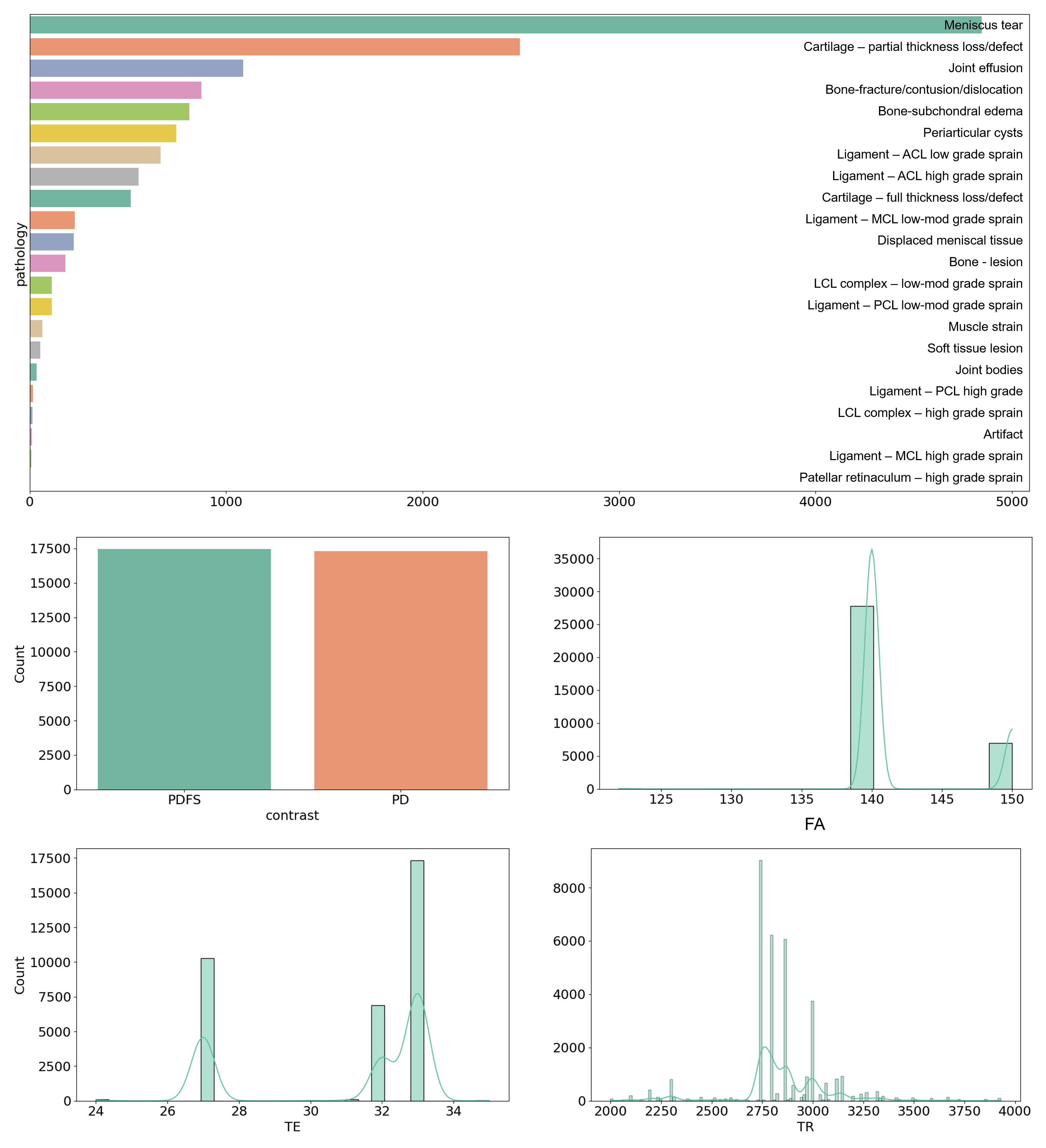}
    \caption{\bf\footnotesize Metadata statistics of fastMRI knee train set}
    \label{fig:metadata_statistic_knee}
\end{figure}

\begin{figure}[!h]
    \centering
    \includegraphics[width=\linewidth]{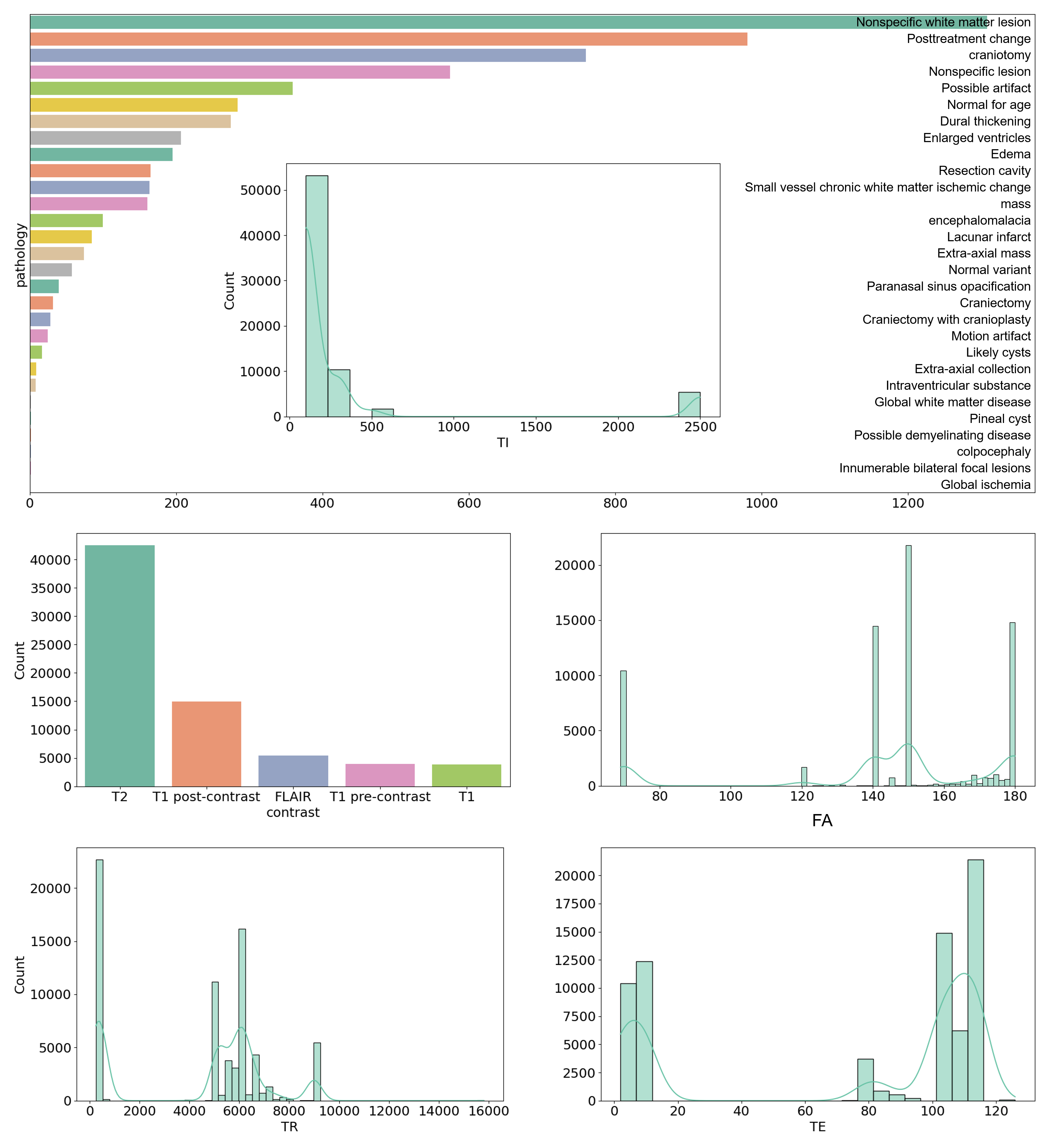}
    \caption{\bf\footnotesize Metadata statistics of fastMRI brain train set}
    \label{fig:metadata_statistic_brain}
\end{figure}

\begin{figure}[!h]
    \centering
    \includegraphics[width=\linewidth]{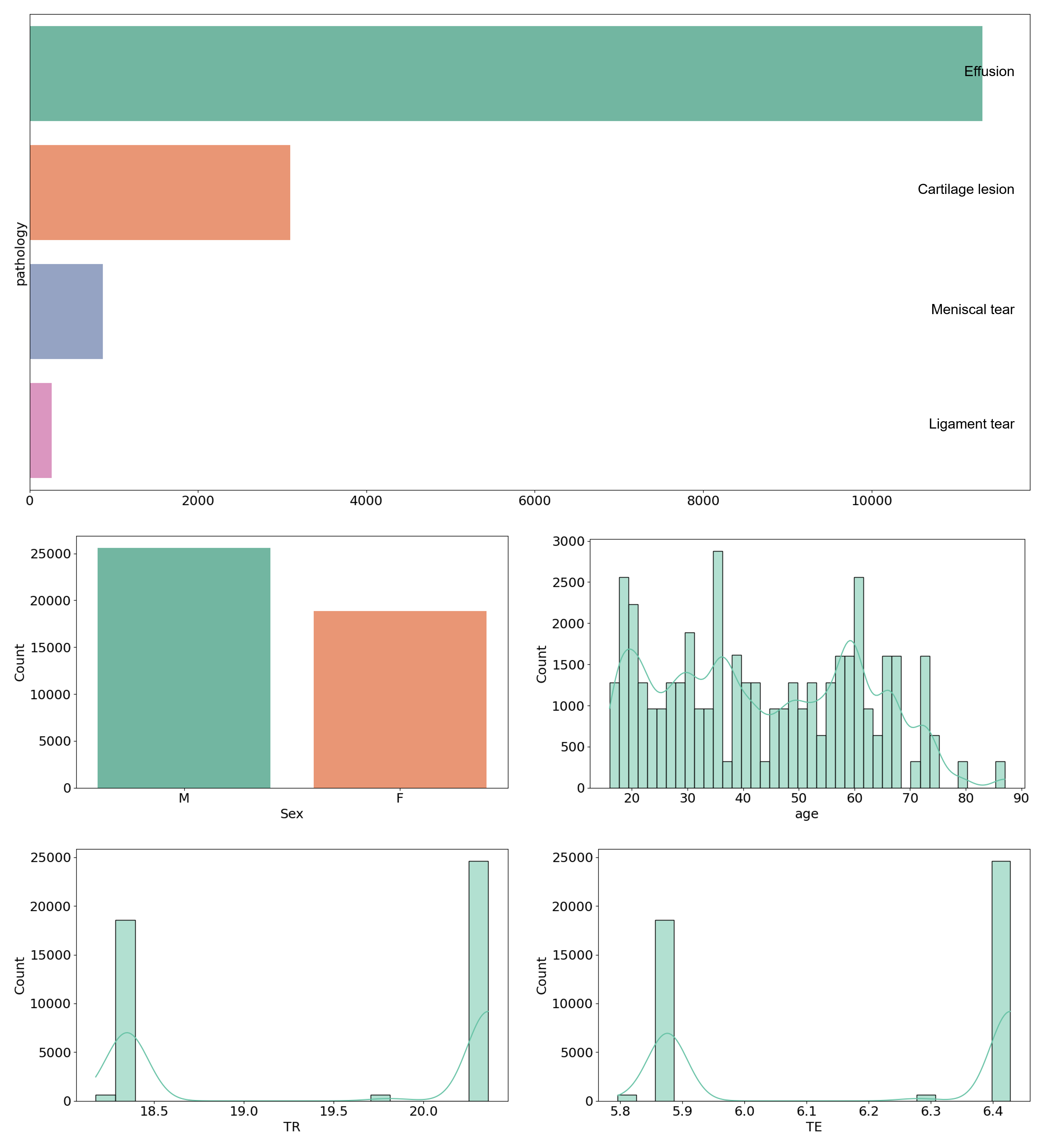}
    \caption{\bf\footnotesize Metadata statistics of SKM-TEA dataset}
    \label{fig:metadata_skm_tea}
\end{figure}

\end{document}